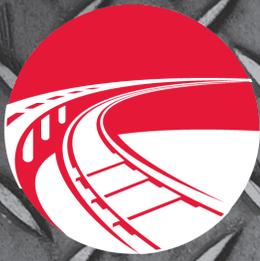

Report # MATC-MS&T: 128-1

Final Report
WBS: 25-1121-0005-128-1

# Modeling and Development of Operation Guidelines for Leader-Follower Autonomous Truck-Mounted Attenuator Vehicles

**Xianbiao Hu, PhD**
Assistant Professor
Department of Civil, Architectural and Environmental Engineering
Missouri University of Science and Technology

**Qing Tang, PhD Student**
Gruaduate Assistant

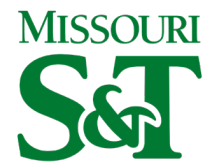

2020

A Cooperative Research Project sponsored by
U.S. Department of Transportation- Office of the Assistant Secretary for Research and Technology

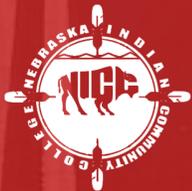

# Modeling and Development of Operation Guidelines for Leader-Follower Autonomous Truck-Mounted Attenuator Vehicles

Xianbiao Hu, Ph.D.
Assistant Professor
Department of Civil, Architectural and Environmental Engineering
Missouri University of Science and Technology

Qing Tang, Ph.D. Student
Department of Civil, Architectural and Environmental Engineering
Missouri University of Science and Technology

# TECHNICAL REPORT DOCUMENTATION PAGE

| 1. Report No. 25-1121-0005-128-1 | 2. Government Accession No. | 3. Recipient's Catalog No. | |
|---|---|---|---|
| 4. Title and Subtitle  Modeling and Development of Operation Guidelines for Leader-Follower Autonomous Truck-Mounted Attenuator Vehicles | | 5. Report Date  December 2020 | |
| | | 6. Performing Organization Code | |
| 7. Author(s)  Xianbiao Hu, PhD, ORCID: https://orcid.org/0000-0002-0149-1847  Qing Tang; | | 8. Performing Organization Report No.  25-1121-0005-128-1 | |
| 9. Performing Organization Name and Address  Mid-America Transportation Center  2200 Vine St.  PO Box 830851  Lincoln, NE 68583-0851 | | 10. Work Unit No. | |
| | | 11. Contract or Grant No.  69A3551747107 | |
| 12. Sponsoring Agency Name and Address  Missouri University of Science and Technology  1201 N State St.  Rolla, MO 65409 | | 13. Type of Report and Period Covered  Final Report (January 2019-September 2020) | |
| | | 14. Sponsoring Agency Code  MATC TRB RiP No. 91994-24 | |
| 15. Supplementary Notes | | | |
| 16. Abstract  Mobile and slow-moving operations, such as striping, sweeping, bridge flushing and pothole patching, are critical for efficient and safe operation of the highway transportation system. A successfully implemented leader-follower autonomous truck mounted attenuators (ATMA) system will eliminate all injuries to DOT employees in follow truck (FT) provided appropriate Statutory authority. The leader-follower system design imposes more requirements to the lead truck (LT) drivers in order to ensure a safe and smooth system operation. The driver is now required to make driving decisions not only from the lead truck's perspective, but also consider the potential implications of his decisions to the follow truck. This project aims to develop a set of rules and clear instructions for ATMA system operation. | | | |
| 17. Key Words  Connected and autonomous Vehicles (CAV); Autonomous Maintenance Technology (AMT); Autonomous Truck Mounted Attenuator (ATMA); Work Zone; Driving Behavior | | 18. Distribution Statement  No restrictions. | |
| 19. Security Classif. (of this report)  Unclassified | 20. Security Classif. (of this page)  Unclassified | 21. No. of Pages  47 | 22. Price |
| Form DOT F 1700.7 (8-72) | | Reproduction of completed page authorized | |

Table of Contents

## List of Figures

# List of Tables

List of Abbreviations

Autonomous Maintenance Technology (AMT)
Autonomous Truck Mounted Attenuator (ATMA)
Automated Driving System (ADS)
Connected and Autonomous Vehicle (CAV)
Cross Track Error (CTE)
Department of Transportation (DOT)
Follower Truck (FT)
Global Positioning System (GPS)
Lead Truck (LT)
Missouri Department of Transportation (MoDOT)
Operator Control Unit (OSU)
Radio Frequency (RF)
Sensitivity Analysis Factor (SAF)
System Control Unit (SCU)
User Interface (UI)
Vehicle-to-Vehicle (V2V)

Disclaimer

The contents of this report reflect the views of the authors, who are responsible for the facts and the accuracy of the information presented herein. This document is disseminated in the interest of information exchange. The report is funded, partially or entirely, by a grant from the U.S. Department of Transportation's University Transportation Centers Program. However, the U.S. Government assumes no liability for the contents or use thereof.

Abstract

Mobile and slow-moving operations, such as striping, sweeping, bridge flushing and pothole patching, are critical for efficient and safe operation of the highway transportation system. Missouri Department of Transportation's (MoDOT) slow moving operations have been crashed into over 80 times since 2013 resulting in many injuries to MoDOT employees. A successfully implemented leader-follower autonomous truck mounted attenuators (ATMA) system will eliminate all injuries to DOT employees in follow truck (FT) provided appropriate Statutory authority. The leader-follower system design imposes more requirements to the lead truck (LT) drivers to ensure a safe and smooth system operation. The driver is now required to make driving decisions not only from the lead truck's perspective, but also consider the potential implications of his decisions to the follow truck. For example, when crossing a highway intersection, a regular driver can simply follow traffic signals and cross intersection at any time in the green phase. However, for an ATMA lead truck driver, he may want to avoid crossing intersections at the end of green phase, otherwise the follow truck won't be able to pass and the lead truck will have to stop and wait for the follow truck to catch up. When scenarios like this happen, it's very likely that the other vehicles will cut in between these two trucks and cause ATMA system failure. Another example is when vehicles are making turns at intersections, an ATMA driver will have to wait for a gap larger than normal to make sure both lead truck and follow truck can make the turns together. This project will develop a set of rules and clear instructions for ATMA system operation.

Chapter 1 Introduction

1.1 Problem statement

Over the last few decades, the states in region 7 have been experiencing major stressors that affect safety performance, with aging infrastructure and lack of maintenance among the top reasons [1]. Mobile and slow-moving operations, such as striping, sweeping, bridge flushing, and pothole patching, are critical for an efficient and safe operation of the highway transportation system. However, the safety concerns associated with those maintenance works are significant and unneglectable. One obvious reason is those maintenance vehicles are operated at a relatively low speed on freeways and highways, while the general traffic is driving very fast. The drastic difference in speed leads to many safety concerns, and if the general traffic fails to switch lanes in a timely fashion, are driving with distractions, or are under the influence, crashes are more likely to happen. On the other hand, research suggests that aggressive or distracted driving does occur very often and is the primary factor in crashes in work zones [2-4]. Other reasons, such as the performance of heavy vehicles [5, 6], speeding [7], and dynamic traffic conditions may also result in work zone crashes [8-10]. Since 2013, Missouri Department of Transportation's (MoDOT) slow moving operations have been crashed into over 80 times resulting in many injuries to MoDOT employees [11]. Such a high number indicates the risks of operating a slow-moving maintenance truck is much higher than driving a regular vehicle, which jeopardizes state DOT employee lives and calls for the need of safer infrastructure maintenance technologies.

Autonomous Truck Mounted Attenuator (ATMA) is a quickly emerging technology [12] and is expected to bring considerable potentials in transportation infrastructure maintenance by removing drivers from risk. The system includes a lead truck (LT), a follow truck (FT), a truck mounted attenuator (TMA) installed on the FT, and a leader-follower system that enables the FT to drive autonomously and follow the LT. The leader-follower autonomous driving system includes actuators, software, electronics, and vehicle to vehicle (V2V) communication equipment that can be installed on TMA-equipped LT and FT. While the LT is performing maintenance work, the FT is designed to serve as the buffer so if a rear-end crash is inevitable, the property damage will be minimized with the TMA hardware installed on the FT. Missouri University of Science and Technology (Missouri S&T) is supporting Micro System, Inc, a wholly owned subsidiary of Kratos Defense (Kratos/MSI) in conducting a leader-follower TMA system study for the Missouri Highways and Transportation Commission (MHTC) and Missouri DOT, to provide a NCHRP 350 level 3 compliant leader-follower TMA system that is capable of operating a driverless rear advanced warning truck in mobile highway operations by 2020, with an ultimate goal of removing MoDOT employees from the follow truck and eliminating injuries while performing slow moving operations.

In order to test the ATMA system's performance, a test event was organized by the Missouri DOT on March 26, 2019 through March 30, 2019, at Fort Walton Beach, Florida, as well as on April 22, 2019 through April 25, 2019, at Sedalia, Missouri. The purpose of the field testing was to test whether the components of the system could meet predefined accuracy and requirements for a minimum of 32 consecutive hours of operation over several days under the controlled environment. GPS data, which is commonly used for a variety of transportation system analysis purposes [13-18], is collected and used in this report to analyze the ATMA system performance. Statistical analysis results suggested that the ATMA system was able to function as expected, and its performance was acceptable when compared with predefined criteria [19, 20]. Moreover, the hypothesis test results suggested that the system was able to function consistently when the testing was repeated, indicating that the system's performance was stable and repeatable.

The field testing was performed under a controlled environment, meaning that the roadway has very limited to no traffic. However, the ATMA system would be deployed in works zones on freeways or highways with variable traffic demands. Therefore, the LT drivers are required to make decisions, not only considering the LT's perspective, but also the potential influence of these decisions on the FT. For example, when making turns at intersections or intending to make lane changes, a general vehicle driver can simply find an acceptable headway gap in the traffic flow and execute his/her desired action. However, the LT driver needs to avoid the situation where the LT can proceed, but the FT has to wait for another acceptable gap to proceed. Under this situation, the LT driver is required to wait for a larger gap in order to make sure that both the LT and FT can proceed with a desired action together. Another example is the car-following gap requirement. According to classic car-following models, the required minimum car-following gap is decided by the operating speed and drivers' response time. However, for the ATMA system, the minimum gap relies more on the accuracy and reliability of the system in maintaining the following distance, but less on the human response time due to the characteristic of autonomous driving.

The research on how DOT engineers should operate the ATMA system on a roadway network is very limited considering those concerns. This research aims to model and develop a set of rules and instructions for ATMA system operators, particularly when it comes to critical locations where decision making is needed. To be specific, technical requirements under three scenarios are investigated: 1) car-following distance, i.e., the minimum safe gap distance between the LT and FT, and that between the LT and the general vehicle in front; 2) critical lane-changing gap distance, i.e., the minimum headway gap for AMTA vehicles, which make sure that both the FT and LT can safely proceed with a desired lane changing action; 3) behavior at an intersection, i.e., the time needed for the ATMA vehicle system to safely cross the intersection. This is important, especially when the vehicles are encountering an ending green light or getting cut off by vehicles with conflicting movements that may discount the autonomous vehicle's capabilities. Traffic flow models are developed in this project, and are calibrated and validated by the data collected from the field testing. The modelling outputs suggest important thresholds for ATMA system operators to follow. It was found that when compared with a common passenger vehicle, these thresholds are significantly higher, which highlights the importance of using the modeling outcomes to train ATMA system operators, as well as providing supplemental work zone traffic management actions to work with the operation of ATMA vehicles to ensure a safe and smooth operation.

1.2 Literature Review

Many car-following models have been developed to illustrate car-following behavior, which describe how a leading vehicle and a following vehicle interact with each other, and is an important consideration to ensure a safe driving experience in a roadway network. The Gazis-Herman-Rothery (GHR) model was first formulated in 1958 at General Motors research laboratory in Detroit [21]. This model related a vehicle's acceleration to the speed of the leader vehicle, relative speed and spacing between the follower and the leader vehicles, and driver reaction time. A safety distance, or collision avoidance (CA) model, was first proposed by Kometani and Sasaki [22]. This model described the safe following distance as a quadratic function of the speeds of the follower and leader vehicles and reaction time, and the four parameters that needed to be calibrated. This model was then further improved by Gipps, in which several mitigating factors were considered [23]. The Gipps model can be calibrated using more common-sense assumptions about driver behavior when compared with the previous CA model. A simplified car-following model was proposed by Newell, in which a follower's trajectory was a simple translation of their leader's trajectory by a specific distance and a time [24]. The relationship between spacing and velocity for a single vehicle was linear related to the specific distance and time. Due to simplicity, the Newell car-following model is applied widely, and it has also been empirically validated in several studies [25-27]. Research on car-following models can also be found in [28-30].

In terms of lane changing (LC) behavior decisions, a number of models were proposed to capture a driver's decision on whether or not to execute an LC. For example, Gipps [31] proposed that a driver's lane changing behavior in an urban street was governed by two basic considerations: maintaining a desired speed and being in the correct lane. Gipps' model considered the driver behavior as deterministic, so that a driver decided to maintain the desired speed or be in the correct lane based on the distance to the intended turn. By extending Gipps model to freeways, Yang and Koutsopoulos [32] classified LC as mandatory or discretionary, and modeled LC as four sequential steps: decision to consider an LC, choice of the target lane, search for an acceptable gap, and execution of lane change. The lead gap was defined as the clear spacing between the front of the lane changer and the rear of the leader in the target lane, and the follow gap was defined as the clear spacing between the rear of the lane changer and the front of the follower in the target lane. The gap acceptance model examines the lead and follow gaps for performing a lane change in the target lane.

Although extensively studied, the above-mentioned car-following and lane-changing models were developed with a single passenger vehicle as their study object and, thus, cannot be directly applied to the ATMA vehicle system. The unique characteristics of the ATMA system, including its LT-FT two-vehicle system design, and autonomous driving capability, as well as the LT-FT distance, all pose additional requirements and require traditional models to be reformulated or modified before being used.

1.3 Research Approach

In this project, traffic flow models are developed to work with the unique ATMA vehicle system design. The Newell simplified car-following model and the classic lane-changing behavior are modified to model the driving behavior of ATMA vehicles at critical decision-making locations. Those two original models are described in following sections.

*1.3.1 Newell Car-Following Model*

A simplified car-following model was proposed by Newell in [24] to describe a passenger vehicle's car-following behavior. This simplified car-following rule assumes that the time-space trajectory of a following vehicle $n$ is essentially the same as the leading vehicle $n-1$, except for a translation in space and in time. Figure 1.1 depicts the characteristics of Newell's car-following model, in which a leading vehicle $n-1$ initially drives at a velocity $v_1$ and then changes to another velocity $v_2$. The actual trajectory should be the dotted line, but for simplification purposes, a piece-wise linear approximation, represented by the solid line, is used. The following vehicle $n$ travels at the velocity $v_1$ at first, and then changes the velocity $v_2$ at a turning point with a spatial delay $d_n$ and a temporal delay $\tau_n$. The delay $d_n$ can be explained as the minimum distance to ensure safe driving and the $\tau_n$ can be explained as the necessary response time of driver $n$.

The following vehicle $n$'s trajectory $x_n(t + \tau_n)$ can be described, as below, in which $x_{n-1}(t)$ is its leading vehicle's trajectory at time $t$, and $\tau_n$ and $d_n$ represent the spatial delay and temporal delay, respectively.

$$x_n(t + \tau_n) = x_{n-1}(t) + d_n \tag{1.1}$$

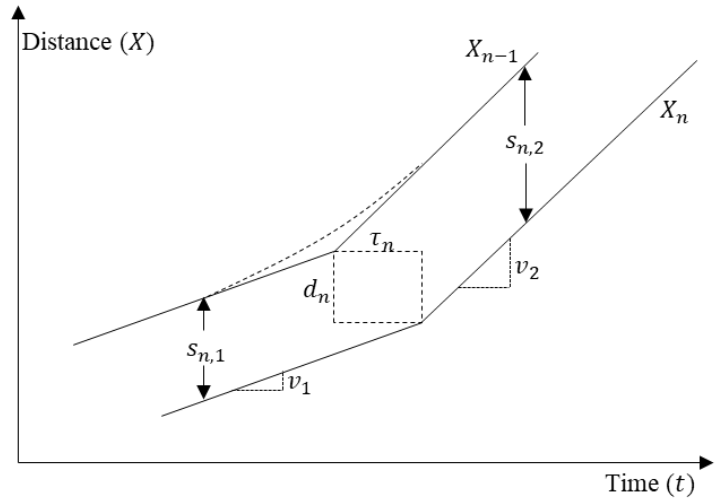

**Figure 1.1** Time-space Diagram of Newell's Simplified Car-following Model

The Newell's car-following model assumes that there exists a linear relationship between the velocity and spacing, as shown in figure 1.2. When the velocity of vehicle $n$ increases, the driver will keep a larger spacing until it reaches the free flow speed. The relationship between spacing $s_{n,i}$ and velocity $v_i$ for vehicle $n$ can be illustrated as below.

$$s_{n,i} = d_n + v_i \cdot \tau_n \tag{1.2}$$

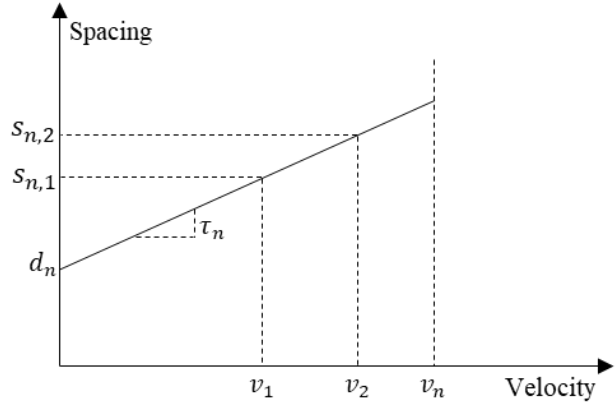

**Figure 1.2** Relationship between Spacing and Velocity for A Single Vehicle

*1.3.2 Critical Gaps for Lane Changing*

Four sequential steps are proposed by Yang and Koutsopoulos [32] to describe a passenger vehicle's lane-changing behavior decisions, including a decision to consider a lane changing, choice of the target lane, search of an acceptable gap, and execution of lane change. Once the target lane is selected, an acceptable gap is required for the driver to change lanes. An acceptable gap consists of a lead gap, lag gap and vehicle length of the subject, as shown in figure 1.3. The subject vehicle at lane 2 intends to change lanes to the target lane 1. The lead gap $L_{\text{lead}}$ is the clear spacing between the front of the subject vehicle and the rear of the lead vehicle in the target lane 1. The lag gap $L_{\text{lag}}$ is the clear spacing between the rear of the subject vehicle and the front of the lag vehicle in the target lane 1. The critical gaps $L$ for lane changing can be expressed as

$$L = L_{\text{lead}} + L_{\text{lag}} \tag{1.3}$$

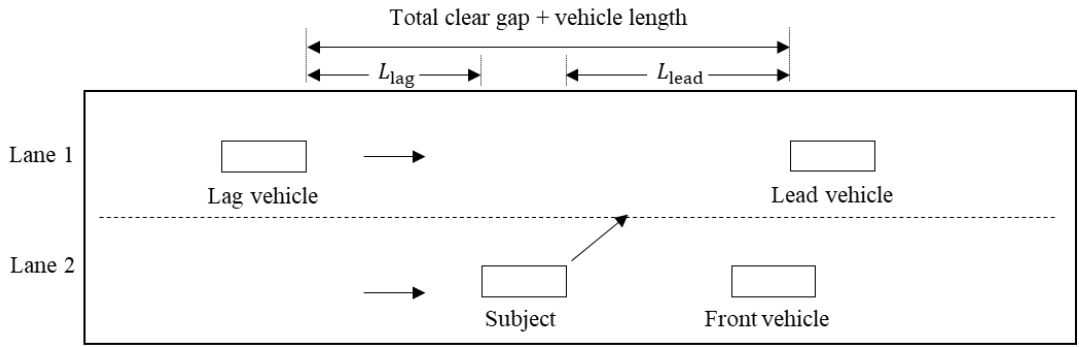

**Figure 1.3** An Illustration of Acceptable Gap for Lane Changing

Due to the characteristics of autonomous driving, human response time is no longer a valid aspect to consider for the FT, rather, the system's capabilities in maintaining a uniform car-following distance and executing lane-changing actions become more important in this process. In addition, the distance between the LT and the FT also plays an important role in the lane-changing process, as we now need both vehicles to switch to a different lane, before the process can be said to be complete. After traffic flow models are developed, the data, collected from real-world field testing and from the sensors installed on an LT and an FT, are used to calibrate and validate the models.

1.4 Organization of the Final Report

Following this introduction, Chapter 2-3 present the overview of ATMA system and field testing. In Chapter 4, the traffic flow models are developed for ATMA system operation. Then the developed models are calibrated and validated using the collected data from the field testing in Chapter 5. The modeling outcomes and implications to ATMA system operations are also presented. Chapter 6 concludes our studies. Literature cited in this report are listed in References.

Chapter 2 ATMA System Overview

The ATMA system operates in a "Leader-Follower" configuration, where the unmanned ATMA vehicle follows behind a human-driven LT that is performing a maintenance operation. The automated driving system (ADS) technology is applied in the "Leader-Follower" system, so that the ATMA system enables manned and unmanned vehicles to perform cooperatively in a multi-vehicle configuration. The ATMA system is shown in figure 2.1. During the ATMA Leader-Follower operation, the software control algorithms and modules capture the movements of the LT, including velocity, heading and position information of the human-driven LT, and transmit the information in packets of data called "e-Crumbs" to the unmanned FT over the Vehicle-to-Vehicle (V2V) Communications link. The transmitted e-Crumbs enable the unmanned FT to follow the precise position, speed, and direction of the LT as it travels along the intended route. A human driver in the LT monitors the performance of the FT and provides backup monitoring of the roadway environment. If the LT changes lanes, moves to avoid an object, etc., the FT will perform the same action at the same position that LT made them. Once the radar detects an obstacle during the performance, the FT will initiate an emergency stop before hitting the obstacle.

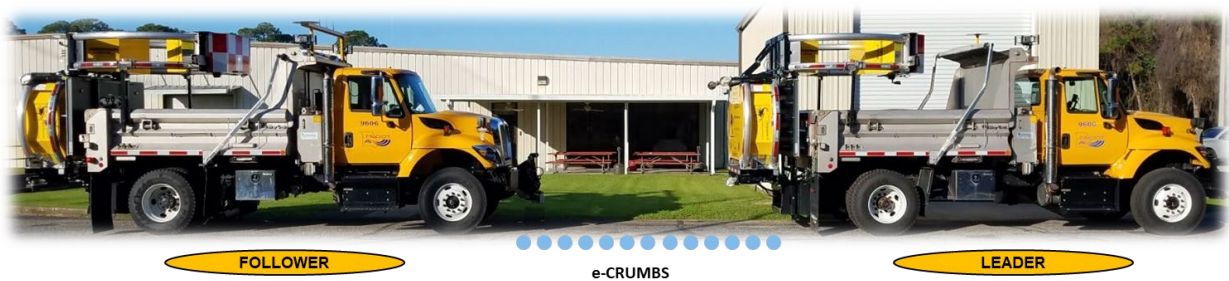

**Figure 2.1** Lead-Follower ATMA Vehicle System

The ATMA system is based on an existing automation kit, originally developed for the U.S. Military, which is the "bolt-on" ATMA Leader-Follower hardware that includes components installed in the LT and driverless FT. The ATMA software control algorithms are optimized for work zone applications and leveraged from countless hours of testing and lessons learned. Similar ATMA systems have successfully supported TMA operations of the Colorado DOT, and Colas United Kingdom, a transportation infrastructure firm. The LT and FT are retrofitted with the components necessary to enable an unmanned operation (with advanced features) that includes: redundancy to eliminate single-point failures, an active safety system, high accuracy Global Positioning System (GPS)/GPS-Denied navigation, encrypted frequency hopping V2V communications, forward-view multi-modal obstacle detect/avoid, side-view obstacle detect/warn, and a robust UI providing system feedback, situational awareness, multi-camera view, and operator controls for vehicle gap adjustment, ATMA pause, a start-up checklist, and offset alignment adjustment.

2.1 Leader Truck Overview

The LT, shown in figure 2.2, is driven by a human driver and performs the roadway maintenance operation in work zones. During the operation, the information of LT, including velocity, heading and position information, are collected by a navigation computer and are transmitted to the LT as "e-Crumbs" through V2C communications link. The performance is achieved by the components and system redundancies on LT shown in figure 2.3 with Backup systems in red and add-on features in blue. These main components are described as follows.

1. **Leader Vehicle System Control Unit (SCU)** serves as the central software component, provides programming and communications technology between the LT and FT, and enables

FT operations. It can back up data simultaneously on multiple channels and protect network, computer, and data from attack, damage, or unauthorized access.

2. **Battery Breaker** powers and protects ATMA components and systems. The design prevents accidental discharge of the vehicle's battery when not in use.

3. **Data radio** enables V2V communication between the LT and FT.

4. **GPS Receiver** provides positioning and velocity data needed for ATMA operations.

5. **Redundant Radio** backup communications links between LT and FT. It works as a backup link between the LT SCU and the FT SCU if primary link goes down. The FT will still be operational if either links goes down.

6. **Operator Control Unit (OSU)** enables the LT driver to start, stop, and monitor LT/FT systems.

7. **Independent E-Stop Initiator and Radio**, which provides the backup capability for the operator to initiate an E-Stop when recognizing an unsafe condition. It provides a failsafe for emergency stop capability that shuts down the automated FT. Loss of power or deterioration of the independent E-stop vehicle communications will automatically trigger an independent E-stop.

8. **I/O Computer and GYRO** are the obstacle detection processing center and backup navigation system in GPS denied circumstances, and it supports navigation during temporary GPS outages.

9. **UI** enables user to conduct and monitor follower operations shown in figure 2.4. It enables gap control and displays system warnings and cautions, including navigation, obstacle detection, side view obstacle detection, live video view of the FT, and the V2V radio communication.

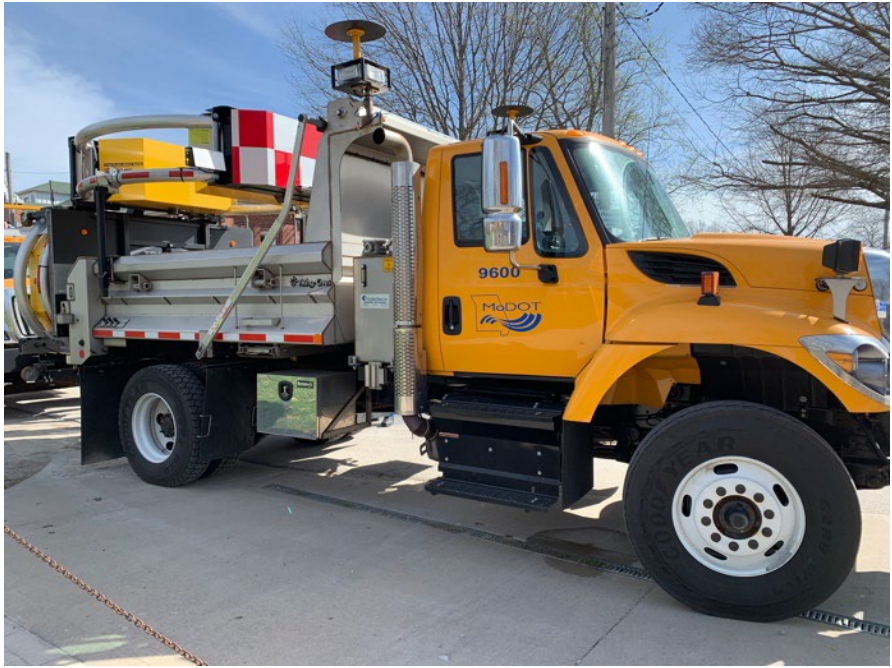

**Figure 2.2** Leader Truck Overview

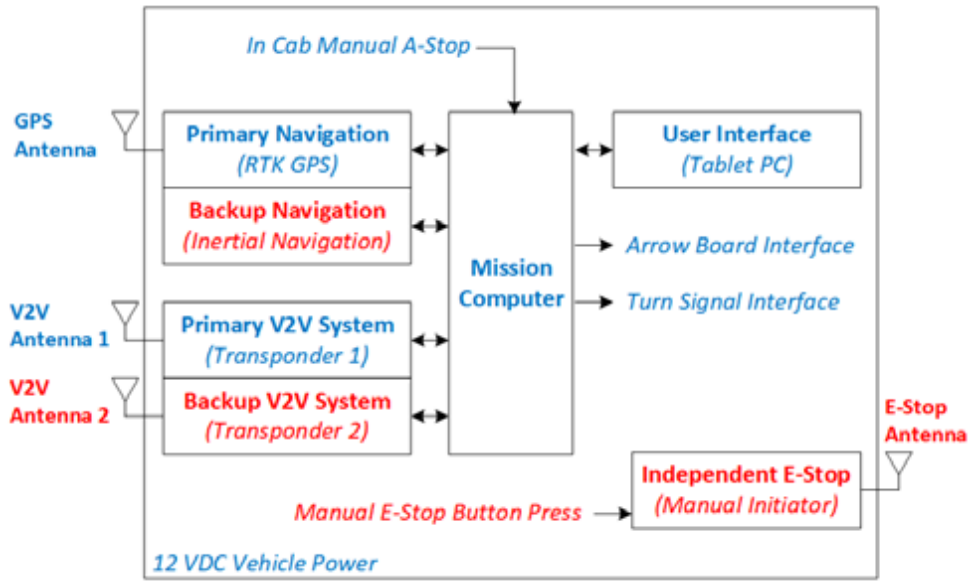

**Figure 2.3** Leader Truck Components

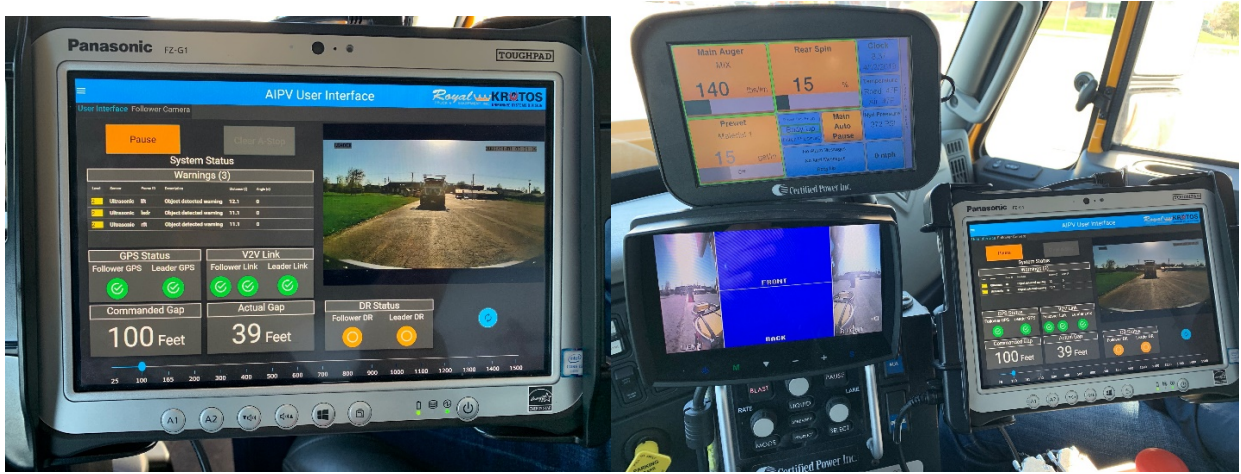

**Figure 2.4** User Interface in Leader Truck

2.2 Follower Truck Overview

    The automated driving system (ADS) technology is applied in the ATMA system, so that the system enables manned and unmanned vehicles to perform cooperatively in a multi-vehicle configuration. During the operation, the FT performs the same movements by receiving the e-Crumbs from the FT, which includes velocity, heading and position information. The transmitted e-Crumbs enable the unmanned FT to follow the precise position, speed, and direction of the LT as it travels along the intended route. The FT is equipped with TMA, which provides protection for human driver in LT presented in figure 2.5. Figure 2.6 shows the mainly equipped components in FT. Different with the LT, the operational safety is provided by the obstacle detection besides the E-stop systems. Apart from the components 1 ~ 8 equipped on LT in Section 2.1, another six main components are installed in FT and are described as follows.

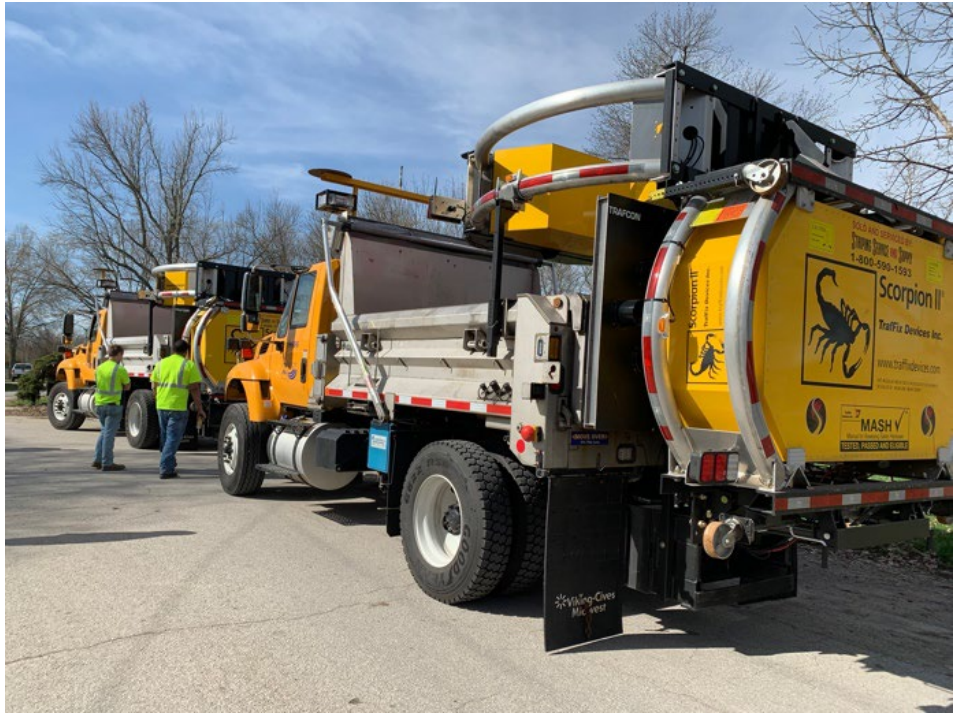

**Figure 2.5** Follower Truck Overview

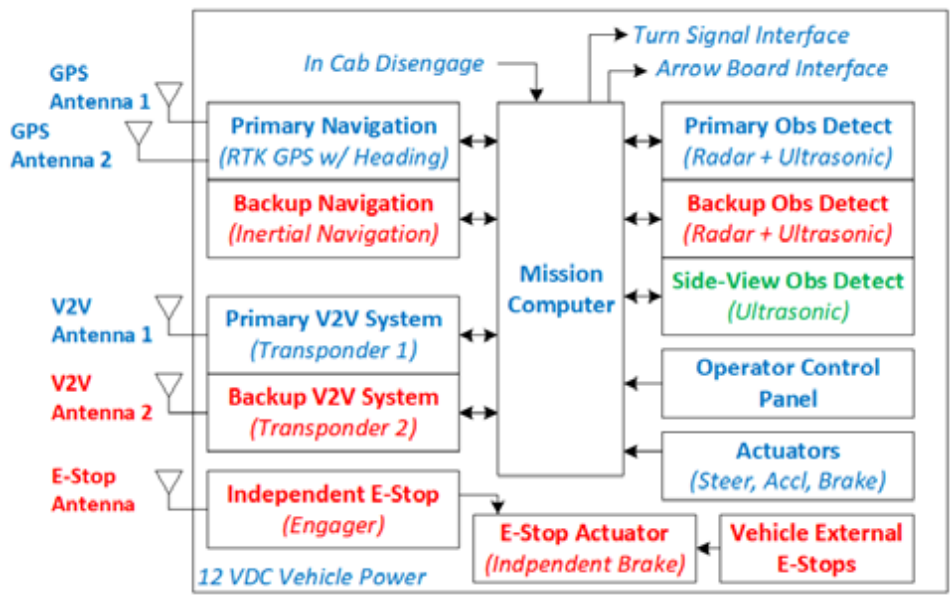

**Figure 2.6** Follower Truck Components

1. **External E-Stop Buttons** allows the worker to stop the vehicle in emergency situations, and it can be installed onto the left, right, and front exterior of the FT.

2. **Front View Obstacle Detection and Avoidance**, which is object detection and avoidance system including Radar, LiDAR, and Ultra Sonic Sensors, and the system will stop the FT if the system is triggered.

3. **Side View Obstacle Detection Warnings,** which consists of four Ultra Sonic Sensors, and thus, can detect and warn the user if there is an object on the side of the FT.

4. **Obstacle Detection and Avoidance Radar** will automatically trigger an A-Stop when an object in the LT path creates an unsafe condition.

5. **Actuator Assembly** can pull or release the cables connected to the break and the accelerator pedals to control the velocity and stop the vehicle.

6. **Steering Actuator** drives the steering wheel to make turns.

Chapter 3 Field Testing Overview

3.1 Time and Location

The dates and locations of the tests were on March 26, 2019 through March 30, 2019 at Fort Walton Beach, Florida, as well as on April 22, 2019 through April 25, 2019 at Sedalia, Missouri.

3.2 Test Cases Overview

A total of 31 cases were defined to test the system's performance. Among them, 23 of the test cases were quantifiable, including 5 on communication loss, 7 on follow distance and accuracy, 3 on obstacle detection, and 8 tests on emergency situations. Each test was repeated for three times to assure statistical accuracy. The other 8 test cases were simple yes/no testing, such as visual inspection on the system and the trucks, data logging, turn signals, and the functionality on the user interface, and are thus not documented in this manuscript.

*3.2.1 Communication Loss*

1. **Simulate Radio Frequency (RF) Loss**. This test was conducted to make sure none-line of sight communication by inserting attenuation (35 Decibel (dB)) into the antenna path to simulate RF loss. After activating LT and FT, the LT drove on curves with a radius of 100 feet at 5 mph. The expected result was that the additional path loss would not cause the lateral tracking accuracy to exceed the specified limit of ± 6 inches. The cross track errors (CTEs) would be recorded to determine whether the test is passed or not.

2. **Loss of Sensor (Radar, LiDAR, Front Facing Ultrasonic)**. This test was conducted to determine if the FT initiated an A-Stop when the Sensor, including Radar, LiDAR, and Front Facing Ultrasonic, was disconnected. The LT and FT drove in a straight line at 10 mph after being activated, and then the technician disconnected one of those three sensors at a time.

The data of distance and time to stop from when Sensor was disconnected were collected to determine whether the FT would initiate an Automatic Stop.

3. **GPS Denied Environment.** This test was designed to test the ability to operate in a GPS-Denied driverless mode with a redundant navigation system. The LT and FT drove in a straight line and through a 100 feet radius turn at 7.5 mph, and then GPS signal of FT was cut. The amount of time that the FT maintained its lane accuracy within ± 6 inches, and the amount time without GPS before the FT initiates an A-Stop were collected. The expected result was that the FT would maintain lane accuracy for a minimum of 45 seconds after GPS was lost, and the FT would initiate A-Stop in under 1 minute.

4. **Loss of Communication (Single V2V Radio).** The test was designed to examine the worst-case lane accuracy in the loss of a single communications channel event. The LT and FT drove in a straight line at 10 mph, and then the technician cut the communications link, one V2V radio, between LT and FT. The worst case lane accuracy in loss of a single communications channel event would be collected, and the UI indication of loss of communications channel event would be recorded. Based on the collected data, the expected result was that the FT would continue to follow the path of the LT without interruption, and the UI would notify the user of the bad communication channel.

5. **Loss of Communication (Both V2V Radio).** This test was designed to determine if the FT would initiate an A-Stop with the loss of both V2C radio communications. The LT and FT drove in a straight line at 10 mph, and the technician cut the communications link, both V2V radios, between LT and FT. The time from loss of communications until A-Stop was initiated, the distance from loss of communications until the FT stopped, and UI indication would be recorded. The expected result was that the FT would execute an emergency stop after the communication was lost.

*3.2.2 Follow Distance and Accuracy*

1. **Following Accuracy on Tangents and Curves.** These two tests were designed to examine the following accuracy on tangents and curves respectively. The FT drove on tangents and curves at a speed of 5 mph, and on curve with a 100 feet radius, which was set up with cones. The worst case lane accuracy on straightaways and while going around curves were recorded. The FT was expected to maintain lane accuracy within ±6 inches from the LT's paths on tangents and curves.

2. **Lane Changing.** This test was designed to examine accuracy during lane changes on two adjacent lanes, marked by cones, with 12 feet wide and 600 feet long. After activating the LT and FT, the right-side lane was closed, and the FT changed lanes from left to right at 5 mph. Then, the FT changed lanes from right to left at 5 mph. The expected result was that the FT would maintain lane accuracy during lane changes.

3. **Bump Test.** This test was designed to examine lane accuracy over minor obstructions in the roadway. In the field test in Missouri DOT, an existing pothole was used as a bump and the FT drove at a speed of 5 mph. The worst case lane accuracy was collected to determine whether the FT would maintain lane accuracy over the minor obstruction in the roadway.

4. **Roundabout.** This test was conducted to determine accuracy during tight turns (roundabout). A roundabout with a 65 feet radius was set up using cones, and the LT and FT drove around it at a speed of 5 mph. The worst accuracy was collected to determine whether the FT would maintain lane accuracy during the tight turns.

5. **Minimum Turn Radius.** This test was conducted to determine accuracy during a 90-degree turn. A 90-degree corner with a 100 feet radius was set up using cones, and the FT turned left and turned right at a speed of 5 mph for three times, respectively. The worst lane accuracy was collected to determine whether the FT would maintain lane accuracy in the turns.

6. **U-Turn.** This test was to determine accuracy during a U-turn. A U-turn path, consisting of a 100 feet straight line and a 65 feet radius, was set up using cones, and the FT made a U-turn from the left and from the right at a speed of 5 mph. The expected result was that the FT would maintain lane accuracy around the turns determined by the worst lane accuracy.

*3.2.3 Obstacle Detection*

The obstacle detection test was conducted to determine if an FT could detect an obstacle that had been placed in its path, and execute an A-stop in time, after the LT passed. Each test was repeated three times for statistical accuracy. A total of three cases were designed and tests were conducted.

1. **Front View Collision Avoidance - Obstacle Detection.** This test was designed to test avoidance of a redundant front view collision. The LT and FT drove in a straight line at 7.5 mph with a gap set at 200 feet. Once the rear of the LT passed the marker barrel at the front of the gap, a technician moved the traffic barrel with a rope. The distance where the FT detected the traffic barrel and the distance between the front of the FT and traffic cone after FT stops were collected. The expected result was that the FT detected the traffic barrel and executed an A-Stop.

2. **Front View Collision Avoidance - Side Obstacle Detection.** This test was designed to test avoidance of a redundant side view collision. The LT and FT drove in a straight line at 7 mph with a gap set at 100 feet. A technician pulled a traffic barrel with a rope, which would be detected at the outside corner of the lane by ultrasonic sensors of the FT. The distance where the FT detected the traffic barrel and the distance between the front of the FT and traffic cone after FT stops were collected. The expected result was that the FT detected the traffic barrel and executed an A-Stop.

3. **Side View Obstacle Detection – Object Recognition.** This case was designed to test redundant side view collision avoidance. In this obstacle detection test case, the LT and FT drove in a straight line at 10 mph. A technician parked a vehicle in the adjacent lane on the left side of the FT. As the FT passed the parked vehicle, the driver looked at the UI for an indication of side collision detection. The expected result was that the object was displayed on the UI.

*3.2.4 Emergency Situations*

Tests of emergency situations were designed to determine the system's performance under emergency conditions, where special operations were needed. A total of eight test cases were defined and conducted.

1. **Temporarily "Drop" the ATMA Vehicle.** This test case was designed to test the ability to temporarily "Drop" the FT and, if the FT would catch up with the LT when the FT initiated a pause command on the UI system. LT and FT drove in a straight line at 10 mph. A technician initiated a pause command on the UI system, to bring the FT to a temporary stop, while the LT kept driving at the same speed in a gap distance of 200 feet for three times. The maximum speed during catch-up and final stabilized gap distance were collected. The expected result was that the FT would catch up to the LT in the set gap distance at a catch-up speed that would not exceed 20 mph.

2. **Emergency Stop.** Four tests were conducted to determine the ability to make an emergency stop for the FT from the LT when executing a stop button, including LT internal button, ATMA internal button, ATMA external button, and LT independent E-stop button. In each test, the FT was expected to stop, and the technician would record the stop time and distance shown.

3. **Braking - Leader Vehicle.** This test was conducted to measure the gap delta between the actual gap and the gap after stopping, when the driver instantly engaged the brake. The gap distance was set to be greater than, or equal to 100 feet. A driver drove the LT at 10 mph, and actual gap was recorded and reported by UI. The driver instantly engaged the brake once the LT passed a limit line marked by cones. The expected result was the FT would stop, and the actual gap between LT and FT shall be recorded.

4. **ATMA Operated by a Human Driver.** This test was conducted to test the take-over capability of an operator in the driver's seat of a FT and the emergency disengagement of an autonomous system. The LT and FT drove in a straight line at 10 mph, and a human driver took control of the FT, after releasing it to the IDLE mode. The expected result was that the FT could quickly disengage from the system, allowing the human driver to take control.

5. **Simulate Rear Impact.** This test case was designed to determine the brake and hazard light functions upon impact. Radar was used three times to simulate a rear impact. The expected result was that the FT would release the throttle, apply full brakes, and turn on the hazard lights.

*3.2.5 Other Tests*

The other 8 test cases were simple yes/no testing, which could be determined whether passed or not directly without furthermore analysis. For example, visual inspection on the system, trucks and UI, data logging, turning signals could be observed after the test.

3.3 Test Data

*3.3.1 Vehicle Operating Mode*

The transition of each vehicle operating mode, from the beginning of a test scenario to the end, was recorded in a log. The log files included three modes for each ATMA vehicle: IDLE, ROLLOUT, and RUN. An IDLE mode is when the safety operator of an ATMA vehicle is in control of a vehicle (instead of the autonomous system having control), and it is the mode in which the ATMA vehicle starts. The FT will transition to the ROLLOUT mode, and then to the RUN mode. The ROLLOUT mode describes the initial state of the ATMA system when operations begin. The initial state is the distance at which the ATMA vehicle travels toward the LT before transitioning to the e-Crumbs navigation method. The travel distance needed to transition to an e-Crumbs path ranges between 20 feet to 30 feet. A ROLLOUT mode should be performed at an approximate speed of 4 miles to 8 miles per hour. The RUN mode is where the FT is operating autonomously.

*3.3.2 Log File Format and Messages*

The FT log file is formatted as a comma separated value (CSV) file format. Figure 3.1 shows a sample of the LT and FT log files.

(a)

| TIMESTAMP | VEH | CRUMB | STAMP | LAT | LON | ALT | HEADING | VELOCITY |
|---|---|---|---|---|---|---|---|---|
| 12:32:35.8 | LDR | 6630 | 18323570 | 38.69311 | -93.261 | -0.024 | 280.004 | 281 |
| 12:32:36.0 | LDR | 6631 | 18323580 | 38.69311 | -93.261 | -0.02 | 279.964 | 279.3 |
| 12:32:36.1 | LDR | 6632 | 18323590 | 38.69311 | -93.261 | -0.022 | 280.004 | 280 |
| 12:32:36.2 | LDR | 6633 | 18323600 | 38.69311 | -93.261 | -0.03 | 279.994 | 280.6 |
| 12:32:36.2 | LDR | 6634 | 18323610 | 38.69311 | -93.261 | -0.016 | 280.004 | 280.2 |
| 12:32:36.4 | LDR | 6635 | 18323620 | 38.69311 | -93.261 | -0.018 | 279.994 | 280.7 |
| 12:32:36.6 | LDR | 6636 | 18323640 | 38.69311 | -93.2611 | -0.028 | 280.104 | 281 |
| 12:32:36.6 | LDR | 6637 | 18323650 | 38.69311 | -93.2611 | -0.026 | 280.164 | 280.1 |
| 12:32:36.8 | LDR | 6638 | 18323660 | 38.69311 | -93.2611 | -0.024 | 280.164 | 280.5 |
| 12:32:36.9 | LDR | 6639 | 18323670 | 38.69311 | -93.2611 | -0.024 | 280.184 | 280.5 |

(b)

| TIMESTAMP | VEH | CRUMB | STAMP | LAT | LON | ALT | HEADING | HDG (Desired) | VELOCITY | VEL (Desired) | GAP | GAP (Desired) | #SATS | VALID | CTE | ACCEL | STEER | STATE |
|---|---|---|---|---|---|---|---|---|---|---|---|---|---|---|---|---|---|---|
| 12:29:08.9 | FLW | 0 | 18290890 | 38.69359 | -93.2615 | 233.84 | 103.473 | 105.085 | 0.01 | 3.97 | 28.48 | 30.5 | 19 | 1 | 0 | -100 | 0 | IDLE |
| 12:29:09.0 | FLW | 0 | 18290900 | 38.69359 | -93.2615 | 233.84 | 103.463 | 105.085 | 0.01 | 3.97 | 28.48 | 30.5 | 19 | 1 | 0 | -100 | 0 | IDLE |
| 12:29:09.1 | FLW | 0 | 18290910 | 38.69359 | -93.2615 | 233.842 | 103.443 | 105.085 | 0.01 | 3.97 | 28.48 | 30.5 | 23 | 1 | 0 | -100 | 0 | IDLE |
| 12:29:09.2 | FLW | 0 | 18290920 | 38.69359 | -93.2615 | 233.842 | 103.523 | 105.085 | 0.01 | 3.97 | 28.48 | 30.5 | 23 | 1 | 0 | -100 | 0 | IDLE |
| 12:29:09.3 | FLW | 0 | 18290930 | 38.69359 | -93.2615 | 233.84 | 103.513 | 105.085 | 0.01 | 3.97 | 28.48 | 30.5 | 19 | 1 | 0 | -100 | 0 | IDLE |
| 12:29:09.4 | FLW | 0 | 18290940 | 38.69359 | -93.2615 | 233.84 | 103.473 | 105.085 | 0.01 | 3.97 | 28.48 | 30.5 | 19 | 1 | 0 | -100 | 0 | IDLE |
| 12:29:09.5 | FLW | 0 | 18290950 | 38.69359 | -93.2615 | 233.838 | 103.483 | 105.085 | 0.01 | 3.97 | 28.48 | 30.5 | 19 | 1 | 0 | -100 | 0 | IDLE |
| 12:29:09.6 | FLW | 0 | 18290960 | 38.69359 | -93.2615 | 233.837 | 103.453 | 105.085 | 0.01 | 3.97 | 28.48 | 30.5 | 19 | 1 | 0 | -100 | 0 | IDLE |
| 12:29:09.7 | FLW | 0 | 18290970 | 38.69359 | -93.2615 | 233.839 | 103.463 | 105.085 | 0.01 | 3.97 | 28.48 | 30.5 | 19 | 1 | 0 | -100 | 0 | IDLE |
| 12:29:09.8 | FLW | 0 | 18290980 | 38.69359 | -93.2615 | 233.838 | 103.473 | 105.085 | 0.01 | 3.97 | 28.48 | 30.5 | 19 | 1 | 0 | -100 | 0 | IDLE |

**Figure 3.1** Screenshot of ATMA Vehicles' Log Files. (a) Leader Truck, (b) Follower Truck

LT log message (see fig. 3.1 (a)) columns are described by the header.

- $1^{st}$ column TIMESTAMP indicates the time in the format hour: minute: second.

- $2^{nd}$ column LCB indicates that it is an LT message.

- $3^{rd}$ column CRUMB indicates the message type as an eCrumb message.

- $4^{th}$ column STAMP indicates the GPS time stamp data for that eCrumb.

- $5^{th}$ column LAT indicates the position in Latitude.

- $6^{th}$ column LON indicates the position in Longitude.

- $7^{th}$ column ALT indicates the Altitude.

- $8^{th}$ column HEADING indicates the heading.

- $9^{th}$ column VELOCITY indicates the velocity of the LT at the eCrumb position in miles per hour.

FT log message (see fig. 3.1 (b)) columns are described by the header.

- $1^{st}$ column is the log timestamp

- 2$^{nd}$ column VEH is the FT message type indicator
- 3$^{rd}$ column CRUMB indicates the eCrumb ID of the eCrumb that the FT is heading towards.
- 4$^{th}$ column STAMP indicates the GPS timestamp for the FT message.
- 5$^{th}$ column LAT indicates the FT's position in Latitude.
- 6$^{th}$ column LON indicates the FT's position in Longitude.
- 7$^{th}$ column ALT indicates the FT's Altitude.
- 8$^{th}$ column HEADING indicates the FT's current heading.
- 9$^{th}$ column HDG(Desired) indicates the FT's desired heading.
- 10$^{th}$ column VELOCITY indicates the FT's current velocity.
- 11$^{th}$ column VEL(Desired) indicates the FT's desired velocity.
- 12$^{th}$ column GAP indicates the FT's current gap.
- 13$^{th}$ column GAP(Desired) indicates the FT's desired gap.
- 14$^{th}$ column #SATS indicates the number of GPS Satellites the Follower is using.
- 15$^{th}$ column VALID indicates if GPS is valid.
- 16$^{th}$ column CTE indicates the FT's cross track error (CTE) which is the horizontal deviation from its intended path and can be obtained from the log files.
- 17$^{th}$ column ACCEL indicates the current brake command.
- 18$^{th}$ column STEER indicates the current steering command.
- 19$^{th}$ column STATE indicates the FT's state (IDLE, ROLLOUT, and RUN).

*3.3.3 Data Processing Procedure*

The log files, generated by the ATMA system, were collected for each test scenario and, during testing, the logged data were exported from the ATMA system once every 24 hours. The vehicle log file was formatted in a CSV format. In some cases, the test data were converted to Keyhole Markup Language (KML) for optional plotting with Google Earth. The log file can be analyzed to gain insights as to the performance and behavior of the system, especially regarding the FT. Data were plotted in Excel, and the statistical characteristics and hypothesis testing were analyzed in Python.

Chapter 4 Model Development for ATMA System Operation

The main purpose of this section is to develop traffic flow models for ATMA vehicle systems, so as quantify its minimum driving requirements. Specifically, technical requirements under three scenarios are investigated, including the minimum required car-following distance for an LT and an FT, respectively, the critical lane-changing gap distance, and the minimum required intersection clearance time. The key to examination of these three scenarios is the modeling of car-following and lane-changing behavior, which is critical to future applications such as travel time prediction [33, 34] and shortest path routing [35-38].

As shown in figure 4.2, a highway with two lanes in each direction and signalized intersection is selected as the analysis target, although the model can be easily extended to other scenarios. It should be noted that Roll Ahead Distance (RAD), which is the distance to allow for forward movement of a vehicle following a rear impact from another vehicles [39, 40], is not modeled in this report, but should be fairly easy to add later on.

4.1 Critical lane-changing gap distance model

The lane changing decision-making process of ATMA vehicles is very different from that of general vehicles. As illustrated in figure 4.1, the ATMA vehicle system consists of an LT and an FT, both of which are located in lane 2 and need to switch to lane 1. Such a leader-follower system design indicates that, the system operator needs to find an acceptable gap for not only the LT, but also for the FT, so that the entire system can make the lane-changing together without being interrupted by general traffic vehicles.

The critical lane-changing gap distance (minimum acceptable gap) for the AMTA vehicles to safely change lanes is denoted as $t_c$. It will include three components, as illustrated in figure 4.1: 1) $L_{\text{lead}}$, which is the required time headway between the LT and the general vehicle ahead in the target lane 1 (i.e., the lead vehicle) so, in case an emergency happens, the LT can come to a safe stop, 2) $L_{\text{lag}}$, which is the required time for the FT to switch lanes, i.e., during this time no other vehicles should cut in between the LT and the FT, and 3) $L_{\text{gap}}$, which is the required time headway between the FT and the general vehicle behind in lane 1 (i.e., the lag vehicle) so, in case an emergence happens, the lag vehicle can also come to a safe stop. The required minimum acceptable gap $L$ can then be calculated as below.

$$L = L_{\text{lead}} + L_{\text{lag}} + L_{\text{gap}} \tag{4.1}$$

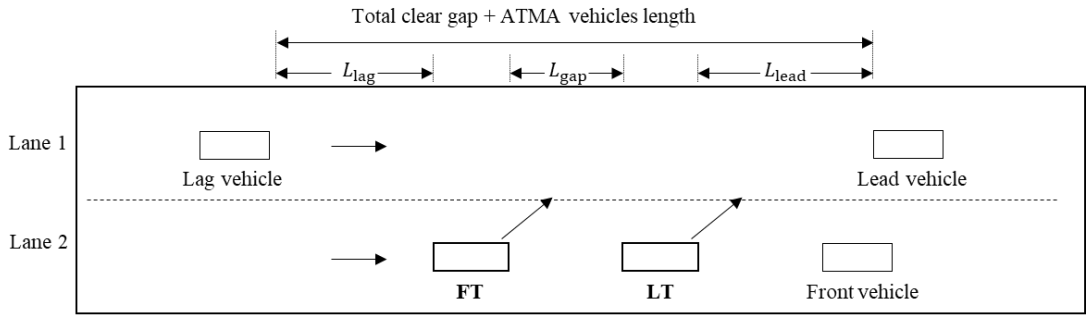

**Figure 4.1** Critical Lane Changing Gap of ATMA Vehicles

Next, we show how to derive $L_{\text{lead}}, L_{\text{lag}}, L_{\text{gap}}$ analytically. Assume, in a general case, a minimum time headway between two vehicles is $t_{\text{min}}$. Note due to crash avoidance requirement, the value of $t_{\text{min}}$ relies on the driving speed $v$, vehicle maximum deceleration $\alpha$ ft/s², and driver response time $t_{\text{rps}}$.

$$t_{\text{min}} = t_{\text{rps}} + v/\alpha \tag{4.2}$$

As such, we can calculate $t_{min}$ for the three above-mentioned scenarios.

$$t_{min,1} = t_{rps,lt} + v_{lt}/\alpha_{lt} \tag{4.3}$$

$$t_{min,2} = L_{gap}/v_{lt} \tag{4.4}$$

$$t_{min,3} = t_{rps,gv} + v_{ffs}/\alpha_{gv} \tag{4.5}$$

where $t_{rps,lt}$ and $t_{rps,gv}$ are the response time of the LT and general vehicle drivers, respectively. Since both vehicles are human-driven, we have $t_{rps,lt} = t_{rps,gv}$. $v_{lt}$ and $\alpha_{lt}$ are the operating speed and maximum deceleration value of ATMA vehicles. $L_{gap}$ is the command gap distance between LT and FT, as after LT makes the lane change, FT will need to travel for a distance of $L_{gap}$ at a speed of $v_{lt}$, before it executes the lane change action. $v_{ffs}$ is the free flow speed of the roadway segment, and $\alpha_{gv}$ is the maximum deceleration value of the general vehicle. The minimum acceptable gap for the AMTA vehicles to safely change lanes $t_c$ can then be calculated by equation (4.6) below.

$$t_c = t_{min,1} + t_{min,2} + t_{min,3} \tag{4.6}$$

4.2 Minimum car-following distance model

*4.2.1 Minimum distance for LT*

In general, the operating speed of a maintenance vehicle is much slower than the other general vehicles, so for the LT, its car-following distance is of less concern while driving in the middle of a roadway segment. However, when the traffic is congested, or when it is approaching an intersection where queuing is observed, the spacing between the LT and a general vehicle ahead becomes shorter and the ATMA system operator needs to watch out for the minimum car-following distance.

According to Newell's simplified car-following model [24], the trajectory of the LT can be expressed as:

$$x_{lt}(t + \tau_{lt}) = x_n(t) + d_{lt} \tag{4.7}$$

Where $x_{lt}$ and $x_n$ are the trajectory of the LT and the general traffic vehicle ahead. $\tau_{lt}$ is the temporal delay which represents necessary reaction time of the driver of the LT. $d_{lt}$ denotes the spatial delay, which stands for the distance that LT traverses from the moment it initiates a brake to the complete stop. According to equation (4.7), the required minimum car-following distance of LT can be expressed as

$$s_{lt} = v_{lt} \cdot \tau_{lt} + d_{lt} \tag{4.8}$$

Where $s_{lt}$ is the minimum required car-following distance, which consists of the distance traveled during the reaction time (the first item on the right) and the distance that the LT travels after a brake is initialed (the second item on the right).

*4.2.2 Minimum distance for FT*

For the FT, due to its autonomous driving nature, the first part of equation (4.8), i.e., the distance traveled during the reaction time equals 0. Once a brake request is initiated, the computer will start to brake without any time delay. As such, when compared with the LT, the required minimum car-following distance only includes the spatial delay $d_{ft}$.

On the other hand, coming into play is the ATMA system's accuracy and stability in keeping the LT-FT gap distance. In order words, if the system operator specifies a command following distance, we also need to consider how accurate the FT can keep up with this desired distance. If we use $\varepsilon$ to denote the 95% percentile of the following distance's error in the FT's car-following distance, the minimum car-following distance of FT can then be derived by

$$s_{\text{ft}} = d_{\text{ft}} + \varepsilon \tag{4.9}$$

With this definition, it can be concluded that as long as the FT keeps a minimum distance of $s_{\text{ft}}$, the chances of the FT following too closely and hitting the LT will become very low, even after considering the potential error of the autonomous driving system.

## 4.3 Intersection clearance time requirement model

In this subsection, we model the minimum clearance time requirement for the ATMA vehicle system at intersections, which is the amount of time needed for vehicles to safely pass an intersection. For a common general vehicle, it merely needs to follow the signal instruction to cross an intersection or to make a turn - as long as it can enter the intersection before the light turns yellow, the designed intersection clearance time is usually sufficient for it go through the intersection, or make a turn. However, for an ATMA vehicle system, this is very different. Considering the LT-FT two-vehicle design, as well as the gap distance between the LT and the FT, the clearance time designed for a single common traffic vehicle is not enough for the ATMA vehicle system. As such, the LT driver needs to assess whether the signal time is sufficient for the ATMA vehicles to cross the intersection.

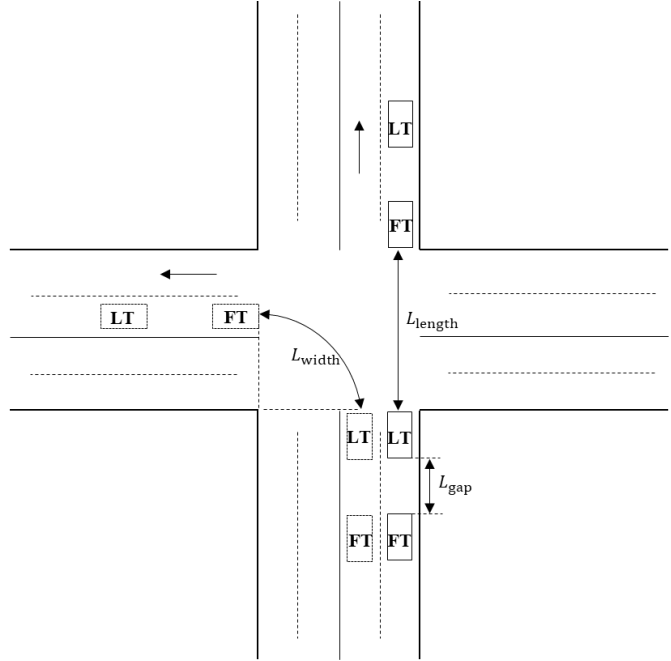

**Figure 4.2** Four-lane Two-way Highway Intersection Profile

The distance of crossing an intersection or making a turn is shown in figure 4.2. If the ATMA vehicles need to cross the intersection, the clearance time requirement for ATMA vehicles can be calculated by

$$t_{\text{straight}} = (L_{\text{length}} + L_{\text{gap}} + 2L_{\text{lt}})/v_{\text{lt}} \qquad (4.10)$$

where the $L_{\text{length}}$ is the length of the intersection, $L_{\text{gap}}$ is the gap distance between two trucks, $L_{\text{lt}}$ is the length of LT or FT, and $v_{\text{lt}}$ is the travel speed of ATMA vehicles. On the other hand, if the ATMA vehicle system needs to make a left-turn at the intersection, the clearance time requirement for ATMA vehicles is calculated by

$$t_{turn} = (L_{width} + L_{gap} + 2L_{lt})/v_{lt} \tag{4.11}$$

where the $L_{width}$ is the width of the intersection.

Chapter 5 Numerical Analysis

In this chapter, we use the data collected from field testing to calibrate and validate the developed models. The modeling outcomes and implications to ATMA system operation are presented in the end.

5.1 Data collection from field testing

In order to derive the minimum required car-following distance and critical lane changing gap, the maximum deceleration of the LT and the FT shall be calibrated. During the field testing, emergency stop testing was performed for three times, which allowed data collection. A technician in the LT pushed the emergency stop button, the ATMA vehicle initiated an emergency stop, and the stop time and distance are shown in table 5.1. The speed of ATMA vehicles was set to be 10 mph and 15 mph.

When the speed is set as 10 mph, the standard deviation (SD) of the stop time is 0.09, whereas the standard deviation of the stop distance is 3.68. When the speed is set to be 15 mph, these values increase to 0.14 and 4.27, respectively. These numbers suggested that the error of the recorded stop distance is larger than that of the stop time. The accuracy of the stop time is considerably higher than that of the stop distance. Because of this, the data of stop time is used to estimate the maximum deceleration. When the speed is 10 mph, the average deceleration is 9.1 ft/s$^2$ and the maximum is 9.4 ft/s$^2$. When the speed is 15 mph, the average deceleration is 11.4 ft/s$^2$ and the maximum is 12.4 ft/s$^2$. In this manuscript, we use the maximum deceleration $\alpha_{lt} = 12.4$ ft/s$^2$ to estimate the car-following distance and the critical lane changing gap.

Table 5.1 The Stop Time and Distance in Emergency Stop Test

| Stop Button | Set GAP | Speed | Stop Time | | | SD | Stop Distance | | | SD |
|---|---|---|---|---|---|---|---|---|---|---|
| | | | Run 1 | Run 2 | Run 3 | | Run 1 | Run 2 | Run 3 | |
| LT Internal | ≥ 100' | 10 mph | 1.56 s | 1.56 s | 1.75 s | 0.09 | 11.5' | 15.75' | 20.5' | 3.68 |
| | | 15 mph | 1.91 s | 2.13 s | 1.78 s | 0.14 | 31.08' | 37' | 26.58' | 4.27 |

5.2 Analysis of car following distance requirement

According to equation (4.8), the minimum required car-following distance of the LT is related to its travel speed and temporal delay. The American Association of State Highway and Transportation Officials (AASHTO) recommend a design criterion of 2.5 seconds for brake reaction time, which exceeds the 90th percentile of reaction time for all drivers [41], as such we set $\tau_{lt} = 2.5s$. $d_{lt}$ is the spatial delay which is the distance that the LT traverses after a brake is applied. The spatial delay can then be calculated by

$$d_{lt} = \frac{v_{lt}^2 - 0}{2 \cdot \alpha_{lt}} = \frac{v_{lt}^2}{24.8} \tag{5.1}$$

According to equation (4.8), the minimum car-following distance of LT can be derived as

$$s_{lt} = d_{lt} + v_{lt} \cdot 2.5 = \frac{v_{lt}^2}{24.8} + 2.5 v_{lt} \tag{5.2}$$

For the car-following distance of the FT, the distribution of errors between the desired gap and the actual gap is plotted in figure 5.1. A positive error means the actual gap is less than the desired gap, which should be included in the minimum car-following distance to ensure safe driving. The 95 percentiles of errors are found to be 6 ft, i.e. $\varepsilon = 6 ft$. According to equation (4.9), the minimum car-following distance of FT can be derived as

$$s_{ft} = d_{ft} + \varepsilon = d_{lt} + 6 = \frac{v_{ft}^2}{24.8} + 6 \tag{5.3}$$

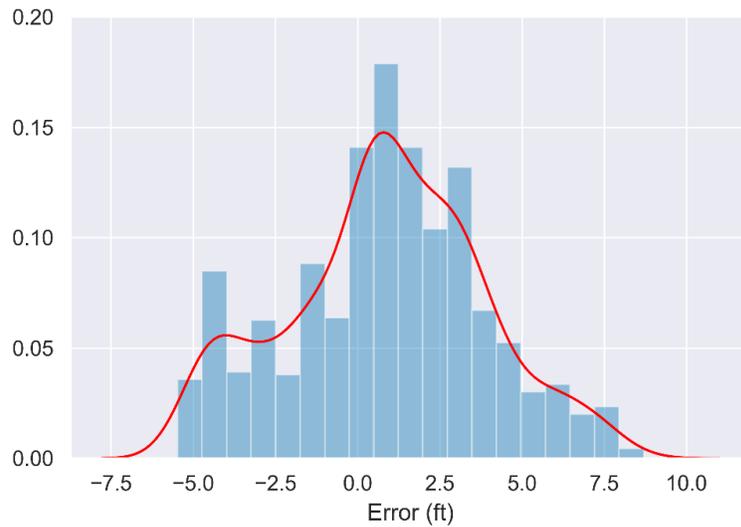

**Figure 5.1** Frequency Distribution Histogram of Error in Follow Distance

Figure 5.2 presents the required car-following distance of LT and FT, which are denoted as $s_{lt}$ and $s_{ft}$, respectively. It can be found that the required car-following distance of LT ranges from 20 ft to 75 ft, and that of the FT ranges from 8 ft to 30 ft. In other words, to ensure a safe driving experience, the required minimum car-following distance should be set to 75 ft for the LT, and a minimum of 30 ft for the FT. It should be noted that, technically, the LT-FT distance can be reduced to as low as 5 ft, but due to safety concerns, a minimum of 30 ft is recommended. In addition, the sensitivity analysis factor (SAF) is adopted to measure the influence of speed change on the car-following distance, which is calculated by

$$\text{SAF} = \frac{\Delta s/s}{\Delta v/v} \tag{5.4}$$

where $\Delta s/s$ is the rate of change of the required car-following distance and $\Delta v/v$ is the rate of change of the operating speed. For the LT, the SAF of the operating speed ranges from 1.13 to 1.27 as the speed increases from 5 mph to 15 mph. For the FT, the SAF ranges from 0.67 to 1.6 as the speed increases. These results indicate that the required car-following distances of LT and FT both have a positive correlation with the operating speed. When the speed increases to over 10 mph, the sensitivity of the FT increases faster than that of the LT.

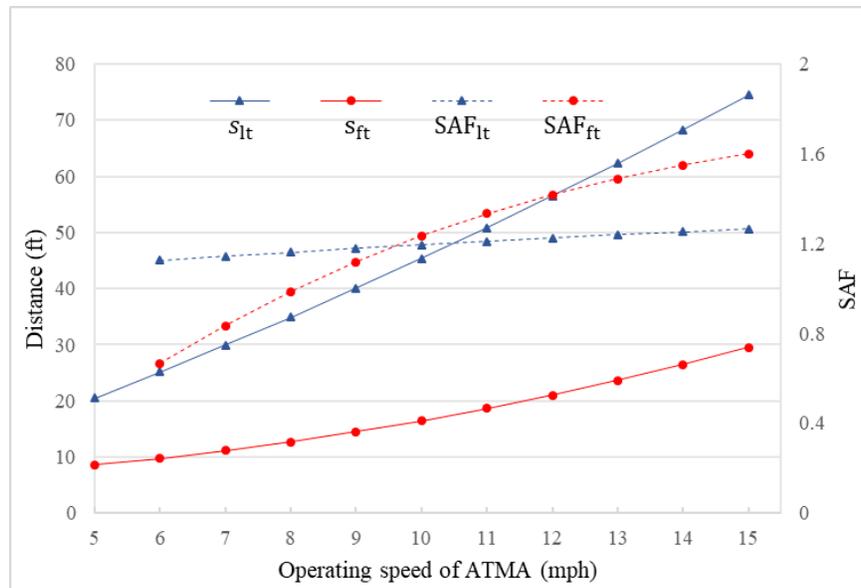

**Figure 5.2** Minimum Car-following Distance of LT

## 5.3 Analysis of Lane Changing Critical Gap Requirement

As discussed in Section 4, the critical gap (minimum acceptable gap $t_{min}$) for the ATMA vehicle includes three components: required time headway $t_{min,1}$ for LT, required time $t_{min,2}$ for FT, and required time headway $t_{min,3}$ for the lagging general vehicle on the target lane.

According to equation (4.3), the time headway $t_{min,1}$ for the LT is related with the operation speed, maximum deceleration, and response time. According to the data from the emergency stop test, the maximum deceleration is 12.4 ft/s². AASHTO recommends a design criterion of 2.5 s for brake reaction time which exceeds the 90th percentile of reaction time for all drivers [41]. As such, $t_{min,1}$ becomes

$$t_{min,1} = 2.5 + \frac{v_{lt}}{12.4} \tag{5.5}$$

The second component, $t_{min,2}$ for FT, is related with the gap distance and the operation speed of ATMA vehicles as illustrated in equation (4.4). In the ATMA system, the gap distance between the LT and FT can be adjusted on a User Interface and ranges between 25' and 1500'. In the field testing this value was set to be 100' to 200', which are the most common in real-world operation and thus we follow such setting to calculate the required time for FT to switch lanes, per equation (4.4). As such, we have $t_{min,2} = \frac{L_{gap}}{v_{lt}}$ in which $L_{gap}$ takes the value of 100' or 200'.

The third component, $t_{min,3}$ for the lagging general vehicle on the target lane, can be derived with free flow speed, maximum deceleration and response time of the general vehicles, as illustrated in equation (4.5). Generally, the free flow speed ranges between 35 mph and 70 mph. Research suggests that approximately 90 percent of all drivers decelerate at rates greater than 11.2 ft/s² (a comfortable deceleration for most drivers), which is recommended as the threshold to determine stopping sight distance [41]. Studies documented in the literature [42] show that most drivers decelerate at a rate less than 14.8 ft/s² when confronted with the need to stop for an unexpected object in the roadway. As such we take 14.8 ft/s² as the threshold to use. $t_{min,3}$ now becomes:

$$t_{min,3} = 2.5 + \frac{v_{ffs}}{14.8} \tag{5.6}$$

In total, the critical gap (minimum acceptable gap $t_{min}$) for the ATMA vehicle to safely change lanes can be calculated by:

$$t_c = t_{min,1} + t_{min,2} + t_{min,3} = 5 + \frac{v_{lt}}{12.4} + \frac{L_{gap}}{v_{lt}} + \frac{v_{ffs}}{14.8} \tag{5.7}$$

Figure 5.3 illustrates the minimum lane-changing critical gap distance, with the LT-FT distance set to be 100 ft in (a) and 200 ft in (b). It can be observed in both cases, when ATMA operation speed increases, the required critical gap reduces, and a higher free flow speed requires a higher critical time headway gap. For a typical freeway with 70 mph FFS, when the gap distance is set to be 100 ft, the critical time headway gap ranges from 18 to 27 seconds, or 23 to 40 seconds with 200 ft LT-FT distance. If we set the ATMA operation speed as 10 mph which is the most commonly-seen, the required critical gap becomes 20 seconds and 26 seconds, respectively. When compared with a common passenger vehicle, these numbers are significantly higher which, again, confirms our previous hypotheses that an ATMA system operator needs to drive a vehicle in a very different way than when driving a common vehicle. Supplemental work zone traffic management actions, such as traffic cones or flaggers, might be helpful to ensure this lane-changing action won't be interrupted by other vehicles.

The SAF of different free flow speeds is also shown in the figure. The value ranges from -0.47 to -0.16 when the gap distance is set to be 100 ft, or -0.6 to -0.35 with 200 ft distance. The negative sign indicates that the critical gap has a negative correlation with the operating speed. We can also find that the absolute of SAF decreases, which means the sensitivity decreases gradually as the operating speed increases. In the meantime, the sensitivity decreases accordingly as the gap distance increases.

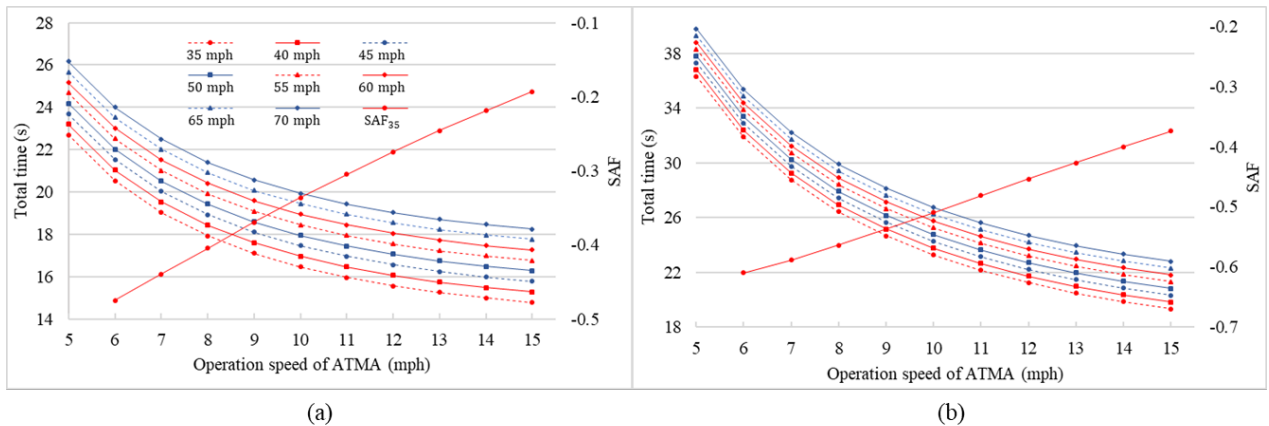

**Figure 5.3** Minimum Lane-changing Critical Gap Requirement: (a) 100 ft Gap Distance; (b) 200 ft Gap Distance

5.4 Analysis of Intersection Clearance Time Requirement

We continue to use the same roadway configuration as specified in figure 4.2. The lane width is set to be 12ft so that the intersection length is 48ft, and the intersection width is $\pi \times (24 + 6) \times 2/4 = 47.1$ ft. The gap distance between LT and FT is also set to be 100 ft and 200 ft. The required time for ATMA vehicles to cross the intersection can be calculated by equation (5.8).

$$t_{min} = (48 + L_{gap} + 2 \times 40)/v_{lt} \tag{5.8}$$

Similarly, the required time for ATMA vehicles to make a left turn is calculated by equation (5.9).

$$t_{min} = (47.1 + L_{gap} + 2 \times 40)/v_{lt} \tag{5.9}$$

Figure 5.4 presents the required time for ATMA to pass the intersection. As the intersection length is about the same with the width, the required time for crossing or making a left turn are almost equal. We can find that the required time for crossing this intersection ranges from 10 to 31 seconds when the gap distance is set to be 100 ft, or 15 to 45 seconds with gap distance of 200 ft. In particular, if the ATMA operation speed is set to 10 mph which is the most commonly-seen, the required intersection clearance time becomes 15 seconds and 25 seconds, for through-movement and a left-turn, respectively. In addition, according to equation (5.4), the SAF of the operating speed ranges from -0.83 to -0.93 when the gap distance is set to be 100 ft or 200 ft, which indicates the required time of going straight or making a turn at the intersection has a negative correlation with the ATMA operating speed. We can also find that the absolute value of SAF increases slightly, which indicates a slight increase of the sensitivity as the ATMA operating speed increases.

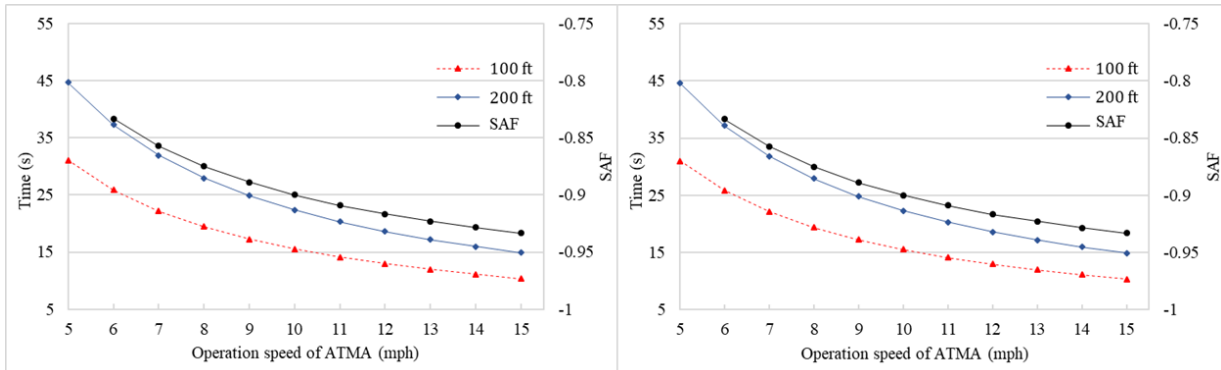

**Figure 5.4** Required Time for ATMA Vehicles at An Intersection to: (a) go straight; (b) turn left.

Chapter 6 Conclusion

This project focuses on modeling and developing a set of rules and instructions to operate the ATMA vehicle system, particularly when it comes to critical locations where correct decision making is needed. Different from general vehicles, the ATMA vehicle system consists of an FT and an LT, with some distance between them. The operators are required to make driving decisions, not only from the LT's perspective, but they must also consider the potential implications of their decisions on the FT. Specifically, three technical requirements are investigated, including those for car-following distance, critical lane-changing gap distance, and intersection clearance time. The Newell car-following model and the classic lane-changing behavior model are modified to model the driving behaviors of the ATMA vehicles at those critical decision-making locations. Data are collected from real-world field testing to calibrate and validate the developed models.

The modelling outputs suggest important thresholds for ATMA system operators to follow, in order to ensure the safe driving of both public and ATMA vehicles. The results suggested a minimum car-following distance of 75 ft for a LT and 30 ft for a FT. In terms of lane-changing critical gap, when the gap distance between the LT and the FT is set to be 100 ft, the system requires a minimum time headway of 20 seconds to perform a safe lane-changing. This number increases to 26 seconds if the FT-LT distance increases to 200 ft. With regard to the intersection clearance time, when the gap distance between the LT and the FT is set to be 100 ft, the system requires 15 seconds to cross an intersection or to safely make a right turn, and this number increases to 25 seconds if the FT-LT distance increases to 200 ft. When compared with a common passenger vehicle, these numbers are significantly higher, which highlights the importance of using the modeling outcomes to train ATMA system operators, as well as to provide supplemental work zone traffic management actions to work with the operation of ATMA vehicles to ensure a safe and smooth operation.